\documentclass[journal,comsoc]{IEEEtran}

\usepackage[T1]{fontenc}
\usepackage{multirow}
\usepackage{diagbox}
\usepackage[caption=true,font=footnotesize]{subfig}
\usepackage{subfloat}
\usepackage{graphics}

\usepackage{ifpdf}

\usepackage{cite}

\ifCLASSINFOpdf
   \usepackage[pdftex]{graphicx}
\else
\fi
\usepackage{amsmath}
\usepackage[cmintegrals]{newtxmath}
\usepackage{array}
\usepackage{url}

\begin{document}
%

\title{Deep Anticipation: Lightweight Intelligent \\Mobile Sensing in IoT by Recurrent Architecture}
%
%
%

\author{Guang~Chen*$^{1}$$^{,}$$^{4}$,
	Shu~Liu*$^{2}$, Kejia~Ren$^{1}$, Zhongnan~Qu$^{2}$, Changhong~Fu$^{5}$, Gereon~Hinz$^{3}$,  and~Alois~Knoll$^{4}$
	\thanks{* indicates equal contribution and corresponding authors}
	\thanks{$^1$ College of Automotive Engineering, Tongji University, Shanghai 201804, China, $^2$  ETH Zurich, $^{3}$  fortiss GmbH, $^{4}$ Technical University of Munich, Faculty of Computer Science, Department of Informatics,   $^{5}$ College of Mechanical Engineering, Tongji University, Shanghai 201804, China}
}

\markboth{Journal of \LaTeX\ Class Files,~Vol.~14, No.~8, August~2015}%
{Shell \MakeLowercase{\textit{et al.}}: Bare Demo of IEEEtran.cls for IEEE Communications Society Journals}
%



\maketitle

\begin{abstract}

Advanced communication technology of IoT era enables a heterogeneous connectivity where mobile devices broadcast information to everything. Previous short-range on-board sensor perception system attached to moblie applications such as robots and vehicles could be transferred to long-range mobile-sensing perception system, which can be used as part of a more extensive intelligent system surveilling real-time state of the environment. 
However, the mobile sensing perception brings new challenges for how to efficiently analyze and intelligently interpret the deluge of IoT data in mission-critical services. In this article, we model the challenges as latency, packet loss and measurement noise which severely deteriorate the reliability and quality of IoT data. We integrate the artificial intelligence into IoT to tackle these challenges. We propose a novel architecture that leverages recurrent neural networks (RNN) and Kalman filtering to anticipate motions and interactions between objects. The basic idea is to learn environment dynamics by recurrent networks. To improve the robustness of IoT communication, we use the idea of Kalman filtering and deploy a prediction and correction step. In this way, the architecture learns to develop a biased belief between prediction and measurement in the different situation. We demonstrate our approach with synthetic and real-world datasets with noise that mimics the challenges of IoT communications. Our method brings a new level of IoT intelligence. It is also lightweight compared to other state-of-the-art convolutional recurrent architecture and is ideally suitable for the resource-limited mobile applications.
\end{abstract}

\begin{IEEEkeywords}
Recurrent Neural Network, Internet of Things, Kalman filtering, Convolutional LSTM, Factor Graph
\end{IEEEkeywords}

\IEEEpeerreviewmaketitle

\section{Introduction}
%
%
%
%
%
%

Imagine in the forthcoming Internet of Things (IoT) era, many objects will have wireless internet access. From the information shared by objects connected to the IoT, an agent can perceive the state of the world. In this case, a more reliable and intelligent surveillance system could be constructed to help prevent mobile applications connected to the IoT, robots and self-driving cars for example, from collision.

\begin{figure*}[!htb]
	\begin{center}
		\includegraphics[width=1.0\linewidth]{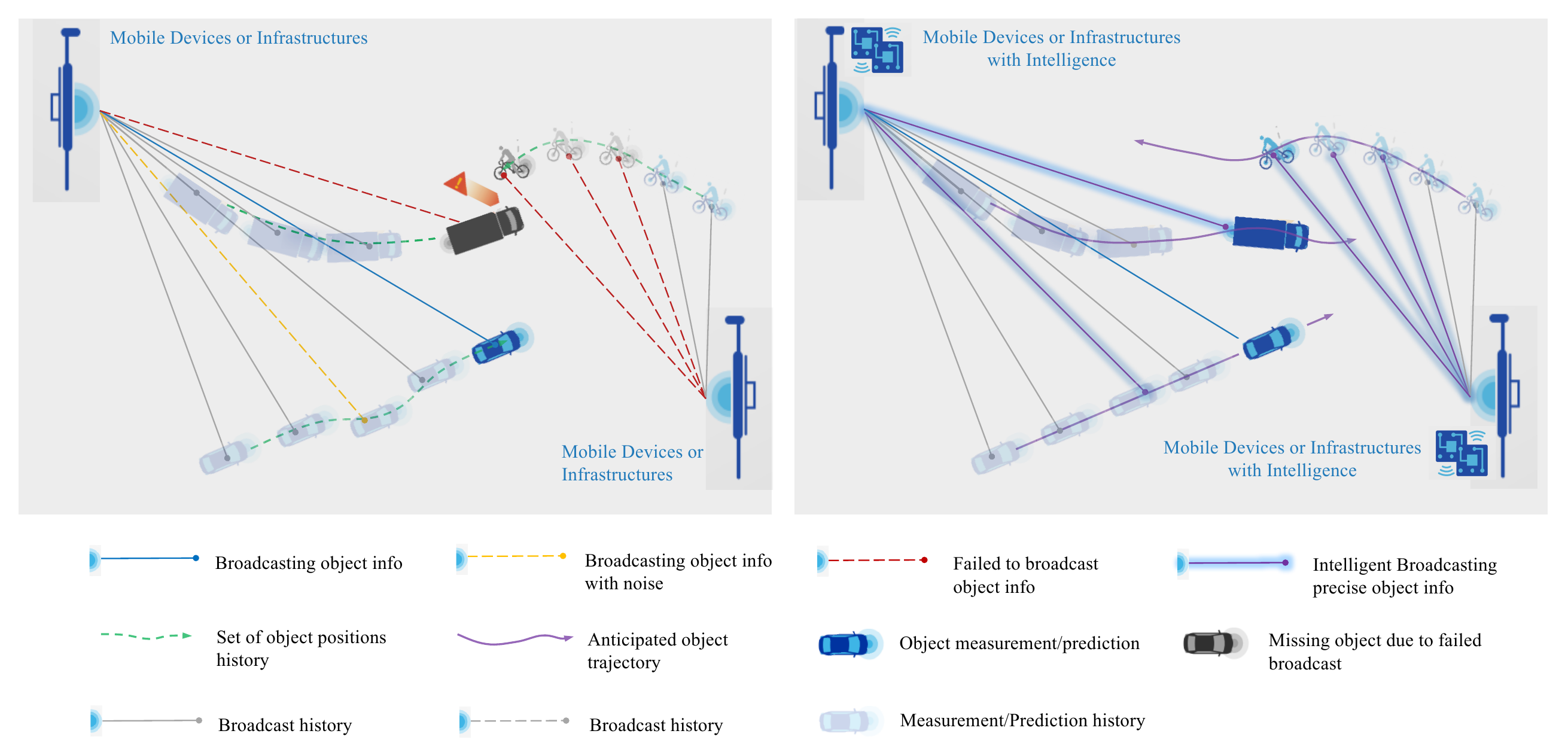}
	\end{center}
	\caption{Illustration of potential problems in IoT based sensing. The vision of this article is to provide a solution targeting these challenges. We leverage RNN and Kalman filtering \cite{kalman1960} to predict motion pattern of partially observable environment and to make correction between observation and prediction.}
	\label{fig:intro}
\end{figure*}

Compared to lidar, radar and camera based sensing, wireless network based perception has several advantages. i) Line-of-sight constraint no longer give a hard limit to the detection range; the broadcast of microwave signal is able to travel around corners and walls etc. ii) The update frequency of wireless networks is potentially much higher than that of lidars and cameras, which are generally limited to $10~Hz$ to $50~Hz$.  In contrast, the upcoming $5G$ network promises a much lower latency of $1~ms$. This property enables many real-time applications that are used in highly dynamic scenarios, such as mobile robots and autonomous vehicles. iii) Through infrastructure or end-to-end communication, mobile devices are able to share information with each other, enhancing the perception area of the environment~\cite{intelligent-iot}. However, the development of IoT based perception brings new challenges for how to efficiently analyze and make sense of the deluge of IoT data in mission-critical services such as autonomous driving and intelligent transportation system. In challenging real world conditions such as crowded or highly reflective environments, wireless communication suffers from high latency, packet loss and decreased throughputs etc. \cite{wirelessnoise}. In such case, the IoT data are unreliable and inaccurate which may lead the mobile sensing perception system to make wrong decisions, e.g. missing vulnerable road user alert at a blind intersection for a V2V system.

To overcome the challenges and harvest the full benefits of IoT data, apart from improving the quality of communication, we propose to combine IoT with rapidly-advancing artificial intelligence technologies to identify and understand the hidden patterns of IoT data. The recent renaissance of artificial neural networks has demonstrated its powerful capability of to deal with spatially and sequentially correlated data. This inspired us to build a intelligent model that is able to infer spatial patterns from sequential IoT data. To incorporate IoT based perception and neural network, we formulate the problem as the following challenges:
\begin{itemize} 
	\item \textbf{Latency}: in crowded or highly reflective environments, the wireless network may become congested, running at high latency. This is critical for real-time application. For example, in autonomous driving, when travelling at the speed of $150~km/h$, a latency of $100~ms$ means the vehicle is partially blind to changes that happen as it travels those $4.17m$. Strongly delayed messages might not be up-to-date any more, resembling packet loss and requiring similar treatment.
	
	\item \textbf{Packet loss}: when communication channels become worse, packet loss could occur due to channel fading or noise corruption. When this happens, sender can be seen as missing from the perspective of receiver. We refer to this kind of noise as \textit{miss noise}. The sender's action can only be anticipated based on its movement history and its current environment, which influences sender's behavior. 
	It should be noted, such a scenario is similar to the occlusion problem in \cite{OndruskaAAAI2016}. However, in our proposed work, we take into account the interaction between objects. 
	\item \textbf{Measurement noise}: Objects participating in the IoT should report their own information and if possible their perception of the environment. Objects can identify their own locations by GPS, IMU, and SLAM etc. Through lidar, radar and camera, objects can also perceive and report objects that are not connected to the IoT. Naturally, all sensors have noise. The locations of objects in a scene may be inaccurate and have shift compared to the ground truth. We refer to this kind of noise as \textit{shift noise}. 
\end{itemize}

In this article, we address the above challenges through a novel combination of a recurrent neural network (RNN) and Kalman-like state \textit{prediction and correction} procedures. This combination of recurrent architecture is able to uncover objects' movement when they are missing from observer's perspective and to recover objects' true trajectories from shift noise.

%
%

\section{Related Work}
Perceiving dynamic environment is one of the most fundamental task for mobile applications. One popular approach to modelling of dynamic scene is to deploy Kalman filtering \cite{TrackingSurvey}. The key idea of this approach is that it assumes measurement inputs and motions are uncertain. Under this assumption, Kalman filtering operates in prediction and correction fashion, namely, prediction of filtering is generated by estimation from dynamical models and afterwards corrected by observation. The belief of the state of a scene is obtained by biased trust of either observation or estimation, depending on their belief distribution. One feature of Kalman filtering is that it relies on hand-designed dynamical models. Therefore, the power of Bayesian filtering is limited to the expressiveness of dynamical models.

To bypass the limitation and burden of hand-designed pipelines, \cite{OndruskaAAAI2016} and \cite{SChoi} frame the tracking problem as a deep learning task. \cite{OndruskaAAAI2016} use recurrent neural network to uncover occluded scene from unoccluded raw sensor inputs, where the network learns an appropriate belief state representation and prediction. \cite{SChoi} implement recurrent flow network, which is able to track objects of different angular velocities. However, most tracking approaches on grid map only consider pepper and salt noise and occlusion. We extend the noise to a more general concept to include inaccurate measurements, i.e. shift of locations of objects. Moreover, both \cite{OndruskaAAAI2016} and \cite{SChoi} do not take interaction between objects into consideration.

Researchers have been exploiting dynamical models and deep learning approaches to tackle interaction of objects in tracking problems. For example, \cite{social-lstm} utilize social pooling and long short term memory (LSTM) \cite{LSTM} architecture to learn spacial-temporal behaviours of pedestrians. However, the method tracks objects individually and may suffer from data association problem in multi-object tracking.

\section{Deep Anticipation}
The occupancy of a scene for applications such as mobile robots or autonomous vehicles is highly dynamic. Objects moving in a scene have interaction with each others. To model occupancy dynamics as a grid map, only the temporal information of an isolated grid cell is not enough. This, in addition to the modelling sequential behaviour of a grid cell,  requires to take information of cells in the vicinity into consideration. 
Moreover, when tracking dynamic occupancy of a scene, the performance can deteriorate if the observation of the scene is highly noisy.

Motivated by the above mentioned challenges,  we build a model that incorporates spatio information into sequential modelling and improves robustness against different kinds of noise. 
In this section, we describe our spatio-pooling strategy and prediction-correction structure for recurrent architecture. In the end, we make a comparison with existing convolutional gated recurrent unit networks. 

\textbf{Problem formulation:}  We model the dynamics of a scene as a occupancy grid map. The advantage of such representation is that \textit{Data Association} of multi-target tracking is avoided. By modelling the states of the cells in grid maps, the dynamics of environments can be obtained. At any time step $t$, we observe the current occupancy of a scene and predict the occupancy of the next time step $t+1$. The prediction is the occupancy probability of each cell. We can tune the threshold level (by default $50\%$) and choose whether to believe a cell is occupied or not.

\subsection{Spatio-Pooling of GRU array}
Gated Recurrent Unit (GRU) networks have been proven to be very effective in learning and presenting sequential data like speech and text \cite{Chung-GRN}. Inspired by this, we use GRU to model the temporal occupancy behaviour of each grid cell. In particular, we assign to each grid cell a GRU cell. That is to say, for a grid map of size $50\times 50$, for example, we deploy also $50\times 50$ GRU cells. In general, we can make the assumption that a scene is homogeneous and the occupancy behaviour is the same everywhere. Under this assumption, in the training phase, we can only train one GRU cell with a batch size of the total number of grid cells in a map, for instance $2500$ in the example above. This design enables faster training thanks to the parallel property and fewer parameters, as compared to convolutional GRU, a GRU version of convolutional LSTM \cite{convlstm}. We refer to this deployment of GRU cells as \textit{GRU array}. It is noteworthy to mention that, we can deploy different GRU cells for a scene, if we assume the scene is inhomogeneous. Though we did not handle such situation in this article, it suggests further improvement in the future. 

Nevertheless, an isolated GRU cell cannot capture enough information to learn and predict dynamics of a scene. The occupancy of a grid cell is also influenced by its neighbours. To address this issue, spatio-pooling of GRU array is used. When modelling the dynamics of a scene, we estimate the occupancy of each grid cell by pooling the hidden states of its own and neighbouring GRU cells. The pooling procedure can be easily achieved using convolutional neural networks. This is different from \cite{social-lstm}, where the authors use RNN to model an individual object (pedestrian) and the tracking is done by pooling the hidden states of other objects in the vicinity. 


\begin{figure}[!htb]
	\begin{center}
		\includegraphics[width=1.0\linewidth]{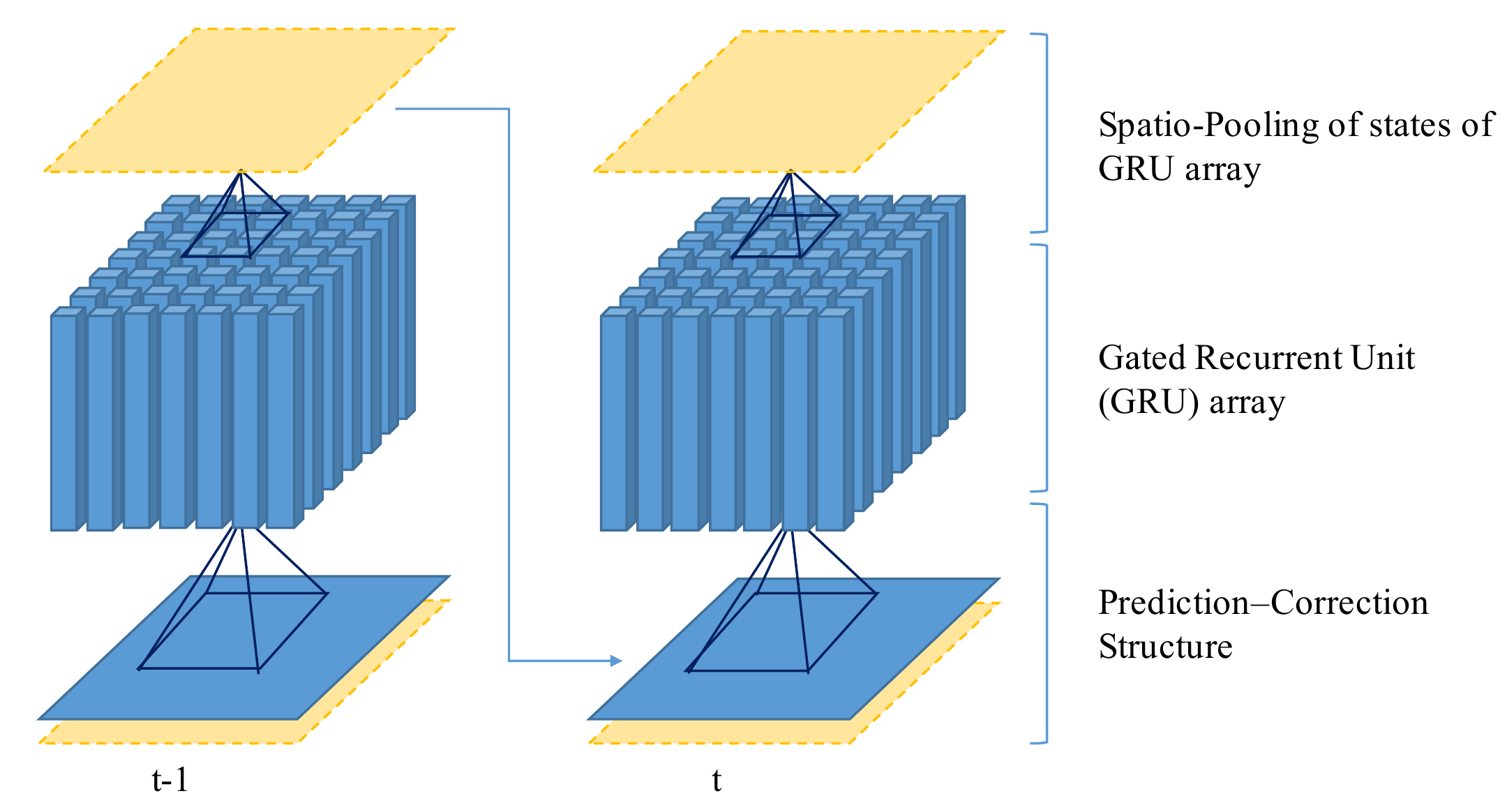}
	\end{center}
	\caption{GRU Array and reuse of previous prediction}
	\label{fig:gruarray}
\end{figure}

\subsection{Prediction and Correction}
\subsubsection{Hidden Markov Model as Factor Graph}
Tracking dynamic scenes or objects can be described by Hidden Markov Models (HMM) \cite{TrackingSurvey}. In \cite{OndruskaAAAI2016}, a graphical model of the generative process of tracking is provided. For the purpose of making it more intuitive, we reframe the model using factor graphs. 


In factor graph, every factor corresponds to a unique node (empty box); each unique variable is represented by an edge or half edge; the node corresponding to some factor $g$ is connected with the edge (or half edge) representing some variable $x$ if and only if $g$ is a function of $x$; variables whose value is constant and known are indicated as solid boxes \cite{sip}.

%

\begin{figure}[!htb]
	\centering
	\subfloat[]{\label{fig:hmm}\includegraphics[width=0.45\textwidth]{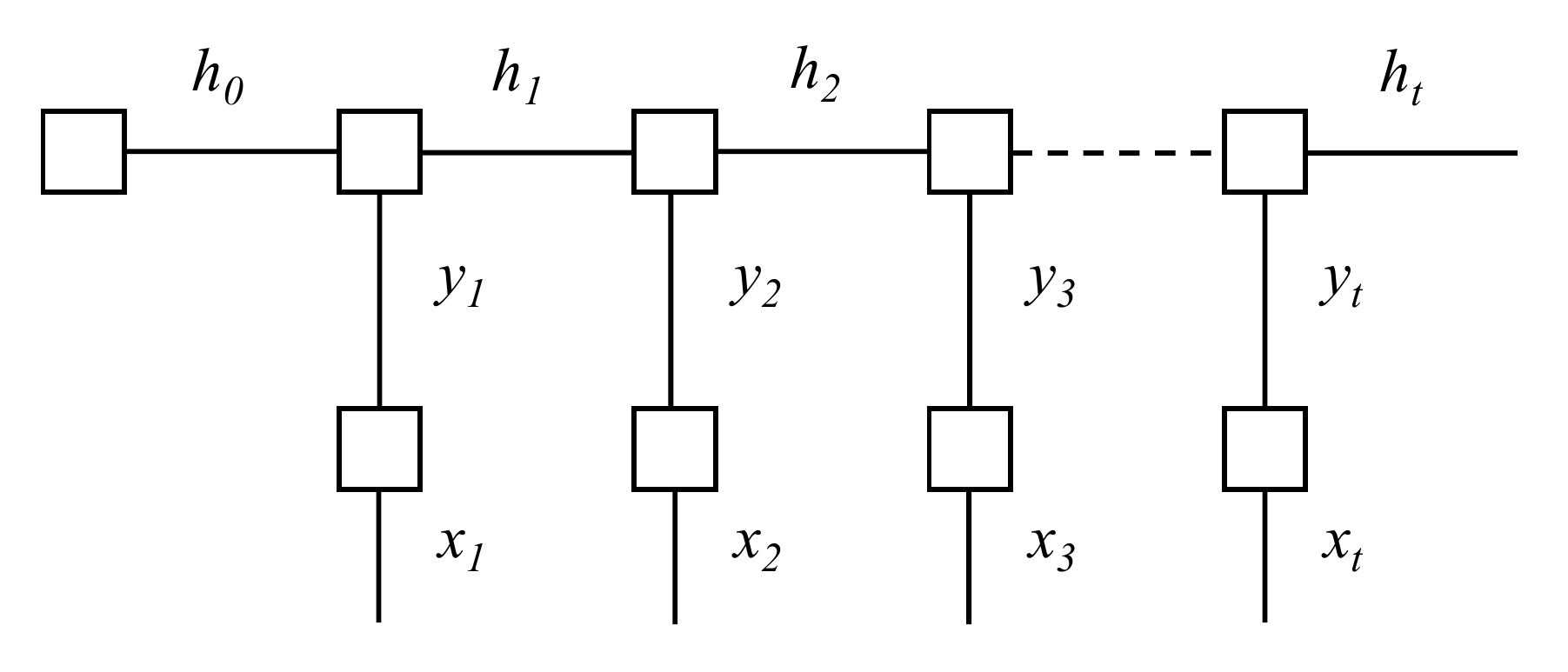}}\hfill~	\subfloat[]{\label{fig:hmm-tracking}\includegraphics[width=0.45\textwidth]{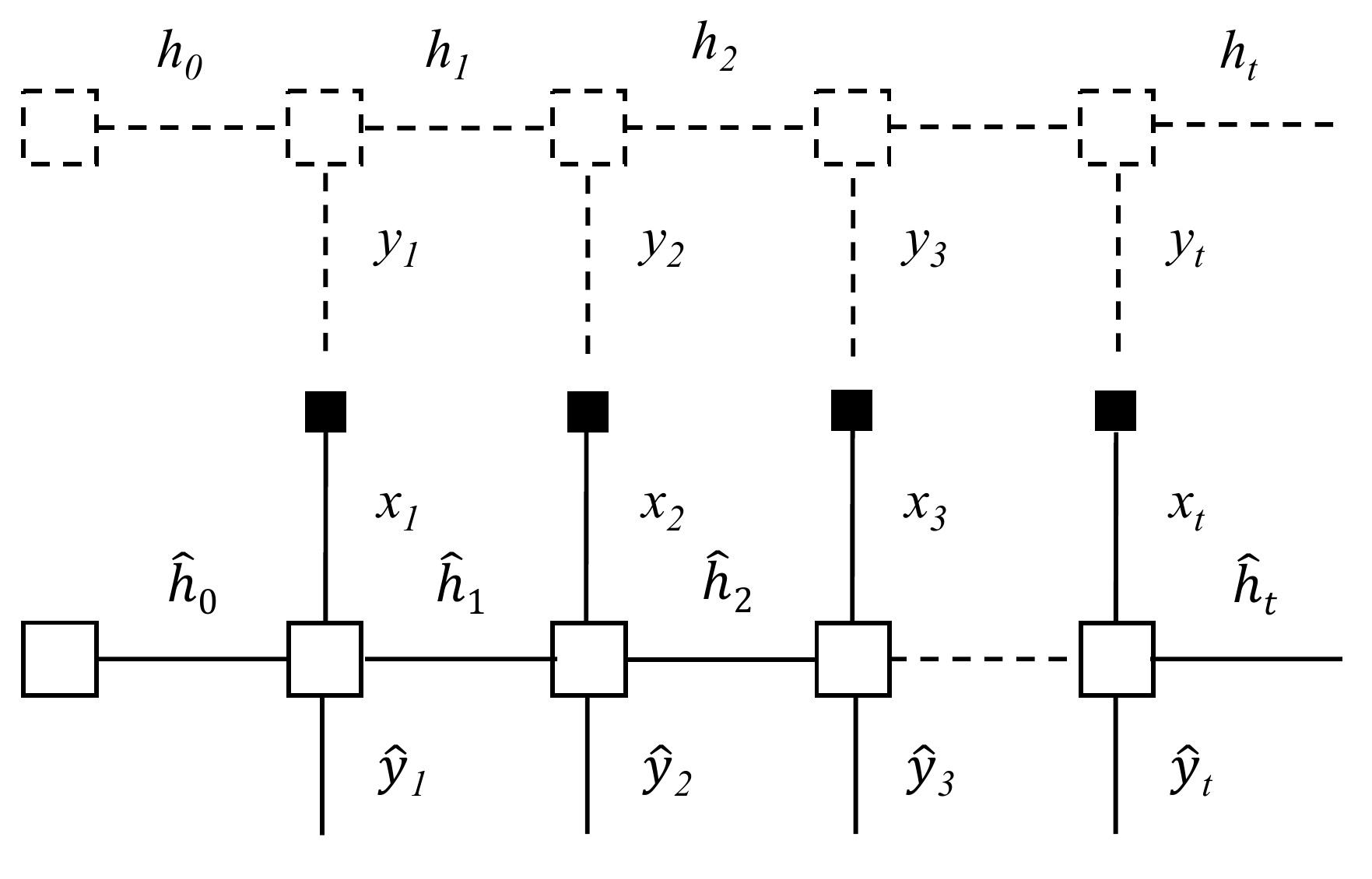}}
	\caption{(a) Hidden Markov Model as Factor Graph. (b) Hidden Markov Model of tracking as Factor Graph.}
	\label{fig:hmm_tracking_ab}
\end{figure}

A factor graph representation of HMM is provided in Fig~\ref{fig:hmm}. Be more specific, the hidden state $h_{t}$ of the model describes true state of the world; $y_{t}$ presents the appearance of a scene such as object positions or occupancy of a map, but does not include the (angular) velocity and acceleration etc., which is necessary for prediction; $x_{t}$ is the perceived information at the receiver and is corrupted by noise. The joint probability density states as follows:

\begin{equation}
\begin{split}
p(x_{0},...x_{t},y_{0}...y_{t},h_{0}...h_{t}) = \\
p(h_{0})\prod_{k=1}^{t} p(x_{k}|y_{k}) p(h_{k}, y_{k}|h_{k-1})
\end{split}
\label{eq:hmm}
\end{equation}

However, from the perspective of tracking system, only the sensor measurement $x_{t}$ are available, based on which the appearance $y_{t}$ and the state of the world $h_{t}$ should be estimated. Usually, the estimation can be done recursively, i.e. prediction of hidden state $\hat{h}_{t-1}$ from the previous step is used for the prediction in the next step. In factor graphs, the situation can be presented as in Fig.~\ref{fig:hmm-tracking}, where only $x_{t}$ are visible to the system. The probability density of the prediction states as follows:

\begin{center}
	\begin{equation}
	\begin{split}
	p(\hat{y}_{t}|\mathbf{x_{1,...,t-1}}) = 
	\int_{\hat{h}_{0},...,\hat{h}_{t-1}} p(\hat{h}_{0}) \\ p(\hat{y}_{t}|\hat{h}_{t-1}) \prod_{k=1}^{t-1}p(\hat{h}_{k}|\hat{h}_{k-1},\mathbf{x_{k}})
	\end{split}
	\label{eq:forwardrnn}
	\end{equation}
\end{center}
where $\mathbf{x_{k}}$ are observations, i.e. known constants. 

\subsubsection{Incorporate Kalman Filtering}
Kalman filtering is a common approach to tracking. It consists of an iterative \textit{prediction-correction} procedure. 
When making prediction, the one-step prediction based on the previous observation is computed; when making correction, the estimate of current state is computed by taking the current measurement into consideration \cite{kalman1960}.

We make use of the idea in our proposed recurrent architecture by recursively making current prediction of hidden state $\hat{h}_{t}$ 
and appearance $\hat{y}_{t}$ dependent on previous prediction of both $\hat{h}_{t-1}$ and $\hat{y}_{t-1}$, i.e. the usage of GRU array and the concatenation of previous prediction with current measurement. The motivation to include $\hat{y}_{t-1}$ in the prediction to improve performance is intuitive. Explained in plain language: if a sensor measurement $\mathbf{x_{t-1}}$ is highly noisy, it is more likely $\mathbf{\hat{y}_{t-1}}$ is closer to the reality; otherwise, $\mathbf{x_{t-1}}$ is more reliable. This recurrent architecture is in line with the underlying principle of Kalman filtering. Mathematically, the probability density of the prediction can be formulated as follows: 
\begin{center}
	\begin{equation}
	\begin{split}
	p(\hat{y}_{t}| \mathbf{\hat{y}_{1,...,t-1}, x_{1,...,t-1}}) =\int_{\hat{h}_0,...,\hat{h}_{t-1}} p(\hat{h}_{0}) \\ p(\hat{y}_{t}|\hat{h}_{t-1})  \prod_{k=1}^{t-1}p(\hat{h}_{k}|\hat{h}_{k-1},\mathbf{\hat{y}_{k}},\mathbf{x_{k}})
	\end{split}
	\label{eq:kalmanrnn1}
	\end{equation}
\end{center}

\begin{figure}[!htb]
	\centering
	\subfloat[]{\label{fig:hmm-prediction}\includegraphics[width=0.32\textwidth]{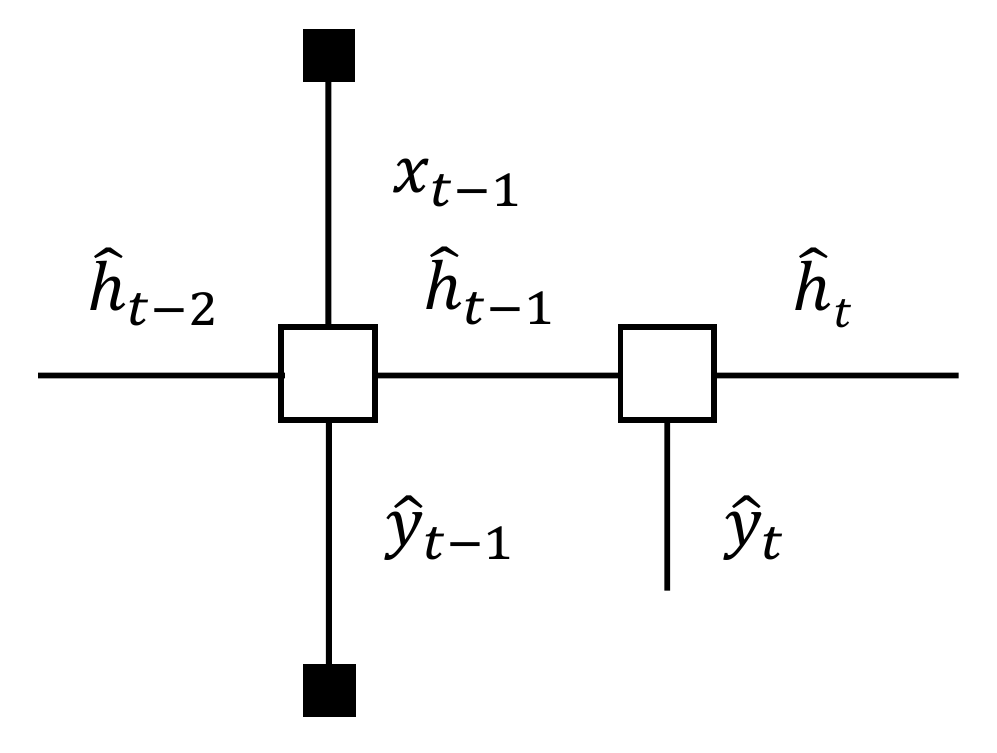}}\hfill~	\subfloat[]{\label{fig:hmm-correction}\includegraphics[width=0.42\textwidth]{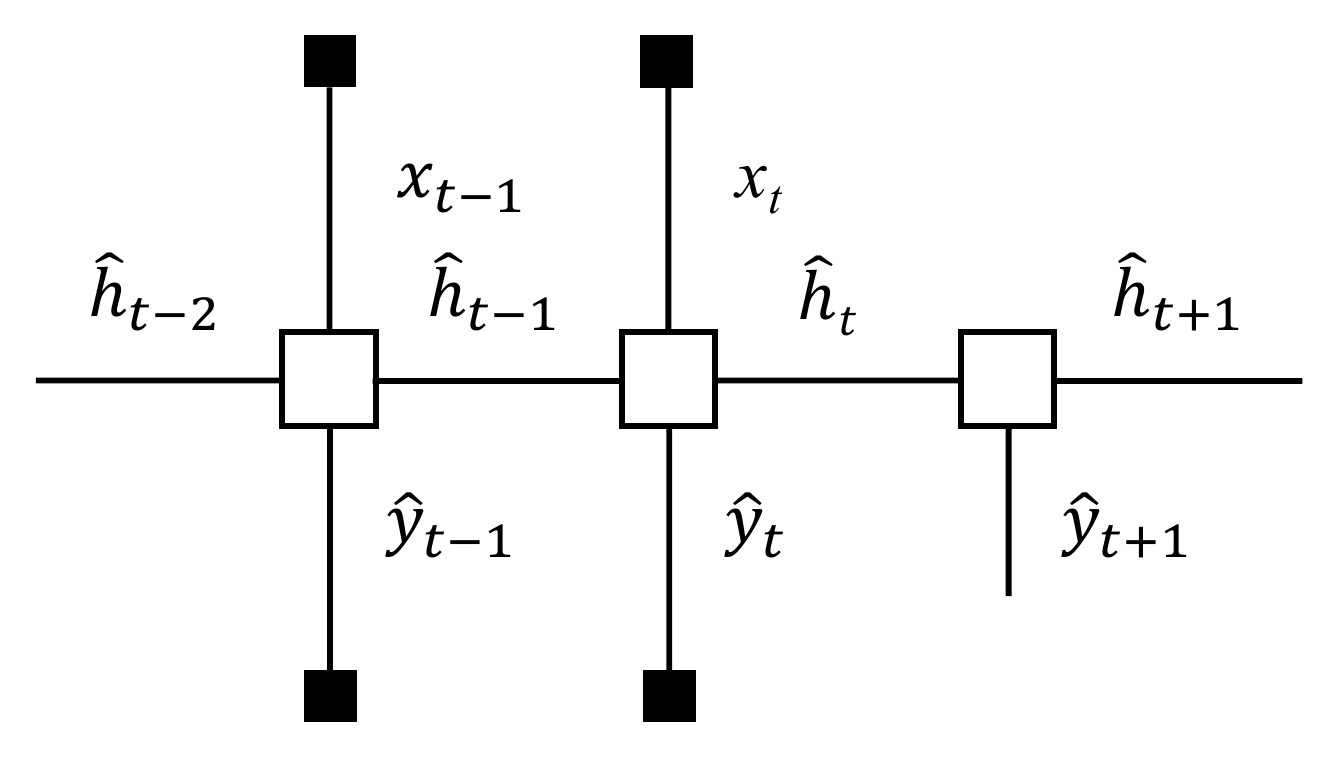}}
	\caption{(a) \textit{Prediction} of $p(\hat{y}_{t}| \mathbf{\hat{y}_{1,...,t-1}, x_{1,...,t-1}}) $. (b) \textit{Correction} using $\mathbf{\hat{y}_{t}} = F(p(\hat{y}_{t}|\mathbf{\hat{y}_{1,...,t-1}, x_{1,...,t-1}}))$.}
	\label{fig:hmm_precor}
\end{figure}

\begin{center}
	\begin{equation}
	\begin{split}
	\mathbf{\hat{y}_{t}} = F(p(\hat{y}_{t}|\mathbf{\hat{y}_{1,...,t-1}, x_{1,...,t-1}}))
	\end{split}
	\label{eq:kalmanrnn2}
	\end{equation}
\end{center}
where $\mathbf{\hat{y}_{k}}$ and $\mathbf{x_{k}}$ are known constants. Moreover, $\mathbf{\hat{y}_{t}}$ is a function $F$ of $p(\hat{y}_{t}|\mathbf{\hat{y}_{1,...,t-1}, x_{1,...,t-1}})$ and $F$ can be anything as long as it leads to a proper prediction from the probability density. The factor graph representation is shown in Fig.~\ref{fig:hmm-prediction} and Fig.~\ref{fig:hmm-correction}. 

%

\subsection{Implementation}
The proposed architecture is named as \textit{Kalman GRN array (KGA)}. We first use 16 convolution filters with kernel size of 6 as encoder to extract spatio information, before passing them as input to the GRU array. The hidden state dimension of each unit in GRU array is set to 16; finally, a convolution filter with kernel size 6 is used to pool hidden states of neighbouring GRU cells and to predict occupancy probability (post-processed by softmax) of each grid cell. For the purpose of visualization, we assign each cell with labels (1 for occupied and 0 for free space) based on probability (threshold by default is 50$\%$); after that, the predicted labels are concatenated with the next measurement as input, as shown in Fig.~\ref{fig:gruarray}. Throughout the whole network, sigmoid is used as activation function; learning rate and optimizer are chosen empirically as 0.003 and RMS-prop; training is terminated through early stopping. For comparison, we build a \textit{ConvGRU} models: GRU array is replaced with convolutional GRU, a variant of convolutional LSTM \cite{convlstm}. A demonstration video is attached with the submission, and also available online\footnote {https://youtu.be/FI0m6IUDvCw}. The source code is available upon request.

\section{Experiment}
\begin{figure*}[!htb]
	\begin{center}
		\includegraphics[width=1.0\linewidth]{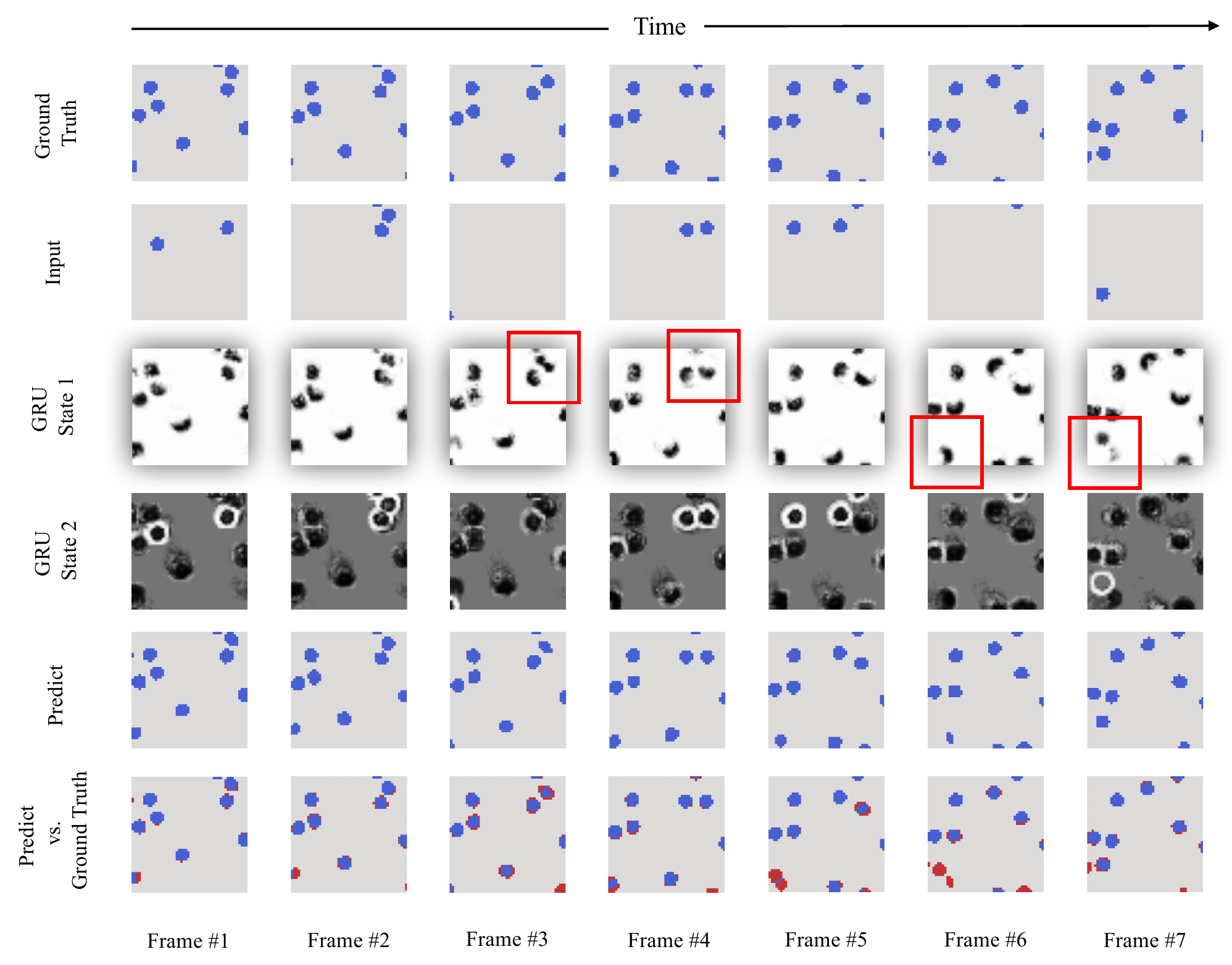}
		\caption{Sensor inputs, activation of states of GRU Array, prediction and ground truth}
		\label{fig:hiddenlayer}
	\end{center}
\end{figure*}

To verify the effectiveness of our model, we present experiments on both synthetic and real datasets. The proposed model is evaluated with binary cross entropy since the correctness of occupancy probability is of our concern. The synthetic dataset is generated using Boids algorithm \cite{boids}, which simulates the flocking behaviour of animals (groups). Moreover, scientists and engineers also applied Boids algorithm for control and stabilization of Unmanned and Ground Vehicles and Micro Aerial Vehicles \cite{boid-robot2}. Because our work focuses on avoidance between individuals, we remove the alignment and cohesion mechanisms and leave only the avoidance function active. 
In addition, two publicly available (human and vehicle) datasets are used for evaluation: UCY \cite{ucy} and NGSIM \cite{lankershim}. In particular, for the NGSIM dataset, we only consider the second segment of the road (intersection) in scene Lankershim. Other segments or scenes (Interstate 80 Freeway and US highway 101) contain mainly straight highways, where vehicles rarely take avoidance or steering action, and thus, the demonstration of anticipation is limited. The UCY dataset contains three scenes: ZARA-01, ZARA-02 and UCY (University of Cyprus). 

Datasets are organized as videos. Each frame is a one channel image, i.e. a matrix of size $50\times50$ binary entries $0$ and $1$ for free and occupied, respectively. Videos are 30 frames per second. Since no specific knowledge about shapes of objects in UCY or NGSIM datasets is given, we use circles with a radius of 2 pixels to indicate pedestrians and with a radius of 3 pixels to indicate vehicles.

The \textit{miss} and \textit{shift noise} are simulated in the following ways. Each object has a certain missing probability (referred as \textit{miss rate}). Namely, some objects are missing in the input frame, and hence, free space occurs where the missing objects should be; 
in addition, for a certain probability, the perceived location of an object may shift from ground truth for up to 2 pixels in both x and y directions (referred as \textit{shift rate}).

Experiments are carried out under conservative conditions. Crucially we set a miss rate to $80\%$ and shift rate to $10\%$. The efficiencies of two models are evaluated on CPU set-up: Intel-i5 2.7 GHz, 8 GB 1867 MHz DDR3.

\begin{figure*}[!htb]
	\begin{center}
		\includegraphics[width=1.0\linewidth]{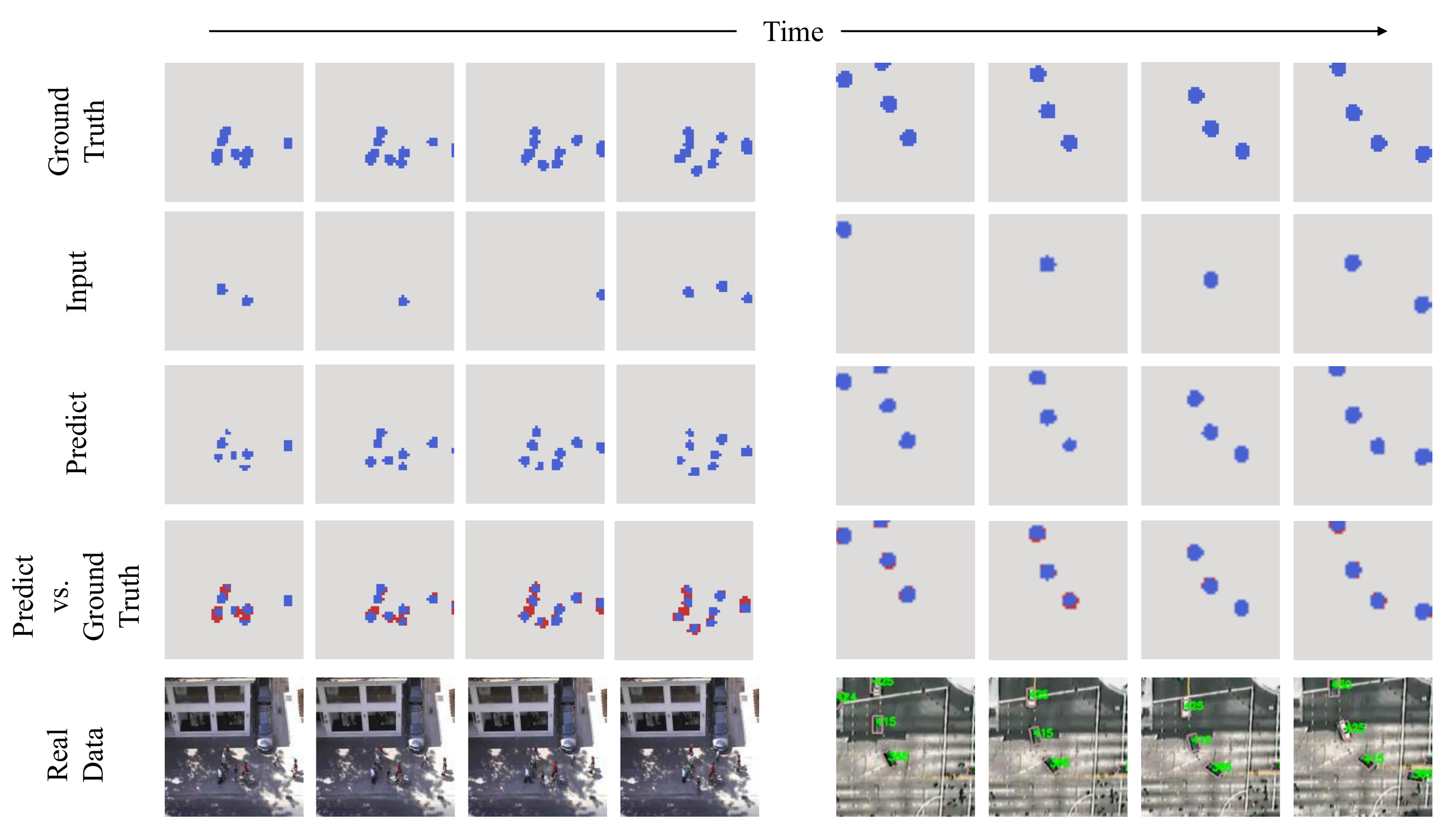}
		\caption{Performance on UCY-dataset (left) and NGSIM-dataset (right). Better view in color version.}
		\label{fig:realdata}
	\end{center}
\end{figure*}

\begin{table*}[!htb]
	\caption{Result on Different Dataset}
	\label{tb:res_all}
	\centering
	\begin{tabular}{|c||c|c||c|c||c|c|}
		\hline
		\multirow{2}*{\diagbox{noise}{model}} &\multicolumn{2}{c||}{Synthetic} & \multicolumn{2}{c||}{UCY} & \multicolumn{2}{c|}{NGSIM}\\
		\cline{2-7}
		& ConvGRU & KGA & ConvGRU & KGA 	& ConvGRU & KGA \\
		\hline
		only miss noise &  \textbf{0.3265}& 0.3306 & 0.3174 & \textbf{0.3172} &  \textbf{0.3188}&  \textbf{0.3188} \\
		\hline
		only shift noise& 0.3235& \textbf{0.3227} & 0.3153& \textbf{0.3149}&  0.3164& \textbf{0.3156} \\
		\hline
		both noises  & \textbf{0.3312} & 0.3351& \textbf{0.3189} & 0.3192 & \textbf{0.3190}&  0.3209\\
		\hline
	\end{tabular}
\end{table*}

An illustration of input and prediction is shown in Fig.~\ref{fig:hiddenlayer} and the supplementary video. While most objects are missing from input frame, the neural network is able to uncover the true dynamics of the scene. In particular, we want to address the activation of hidden states in GRU array. For this purpose, two GRU array hidden states that demonstrate anticipating ability are plotted. First, one can easily notice that, in \textit{state 1}, the moon shape patterns correspond to the motion directions of each object. \textit{State 2} predicts current occupation of the grid, meanwhile its highlight parts coincide with the input measurements. This means the neural network treats prediction and observation differently. Both states memorize object locations and movements, while cooperatively predicting and correcting based on observations.

We describe two concrete examples. First example, in the upper right corner of \textit{state 1} at frame \#3, as marked with bounding box, when an object is about to collide with the other, the moon shape pattern splits into two halves, corresponding to two possible moving directions for avoidance; in frame \#4 however, a new measurement is obtained, as shown in the \textit{Input} and \textit{state 2} rows, the network updates its states and the correct moving direction is computed, as shown in the bounding box of \textit{state 1} at frame \#4. Second example, the object in the bottom left corner of \textit{state 1} at frame \#6  is predicted based on single incomplete observation (in the bottom left of \textit{Input} at frame \#3); however this prediction (moving rightwards) is incorrect (truth is moving upwards); at frame \#7, a new measurement is obtained, as shown in \textit{Input} and \textit{state 2}, the network update its state and the wrong state fades away. An illustration of prediction performance of KGA in real dataset is provided in Fig.~\ref{fig:realdata}. 

The quantitative results are listed in Tab.~\ref{tb:res_all}. Overall, KGA achieves comparable performance as ConvGRU. Moreover, the total number of trainable parameters of KGA are only 3906, while ConvGRU requires 30626. In CPU set-ups, the process speed of KGA is about $5~ms$ per frame while ConvGRU needs about $18~ms$ per frame. This lightweight model enables mobile applications on embedded devices and also makes potential online training strategy possible.


\section{Conclusion}
In this article, we introduced the Kalman GRU array (KGA), a recurrent neural architecture that leverages RNN and Kalman filtering. Moreover, we have presented a promising approach for intelligent mobile sensing in the IoT and the proposed KGA model can anticipate the motions of interacting objects, which in the future could be used for intelligent surveillance systems to help avoid potential traffic collisions. The KGA achieved comparable performance with state-of-the-art methods on both synthetic and real datasets, while using only about $10\%$ of parameters. The computation time is 3 times faster than the state-of-the-art convolutional GRU, which enables lightweight usage on mobile devices. In future work, we plan to explore the possibility of unsupervised training methods. In addition, KGA can be applied to other prediction tasks, such as human action recognition, video stabilization or image generation, where data is spatio-temporally correlated.

%

\section*{Acknowledgment}
The authors would like to thank Federal Ministry of Transport and Digital Infrastructure of Germany for partially funding this project in the frame of Project Providentia.


\ifCLASSOPTIONcaptionsoff
  \newpage
\fi




\small{
	\bibliographystyle{./bibtex/IEEEtran}
	\bibliography{./bibtex/IEEEabrv,IEEEabrv}}
\end{document}